\documentclass[lettersize,journal]{IEEEtran}
\usepackage{amsmath,amsfonts}
\usepackage{algorithmic}
\usepackage{algorithm}
\usepackage{array}
\usepackage[caption=false,font=normalsize,labelfont=sf,textfont=sf]{subfig}
\usepackage{textcomp}
\usepackage{stfloats}
\usepackage{url}
\usepackage{verbatim}
\usepackage{graphicx}
\usepackage{cite}
\usepackage[colorlinks=true]{hyperref}
\usepackage{cleveref}
\usepackage{xspace}
\usepackage[numbers]{natbib}
\usepackage{booktabs}
\usepackage{multirow}
\usepackage{pifont}
\newcommand{\cmark}{\ding{51}}
\newcommand{\xmark}{\ding{55}}

\providecommand{\methodabbr}{Uni-ISP\xspace}
\providecommand{\methodfullname}{\textbf{D}evice-\textbf{a}ware \textbf{ISP}}
\providecommand{\methodfullnamebold}{\textbf{D}evice-\textbf{a}ware \textbf{ISP}}
\providecommand{\gfullname}{inverse ISP module\xspace}
\providecommand{\hfullname}{forward ISP module\xspace}

\begin{document}

\title{\methodabbr: Toward Unifying the Learning of ISPs from Multiple Mobile Cameras}

\author{Lingen~Li,
        Mingde~Yao,
        Xingyu~Meng,
        Muquan~Yu,
        Tianfan~Xue,~\IEEEmembership{Member,~IEEE,}
        and~Jinwei~Gu,~\IEEEmembership{Senior Member,~IEEE}
\thanks{Lingen Li, Mingde Yao, Xingyu Meng, Muquan Yu, Tianfan Xue, and Jinwei Gu are with the Chinese University of Hong Kong (lgli@link.cuhk.edu.hk; mingdeyao@cuhk.edu.hk; stephenmang@link.cuhk.edu.hk; mqyu@link.cuhk.edu.hk; tfxue@ie.cuhk.edu.hk; jwgu@cuhk.edu.hk).  
Personal use of this preprint material is permitted.
Permission from IEEE must be obtained for all other uses. \\Project page: \url{https://lg-li.github.io/project/uni-isp}.
} %
}

\markboth{}%
{Li \MakeLowercase{\textit{et al.}}: Uni-ISP: Toward Unifying the Learning of ISPs from Multiple Mobile Cameras}

\maketitle

\begin{figure*}[t]
  \centering
   \includegraphics[width=\linewidth]{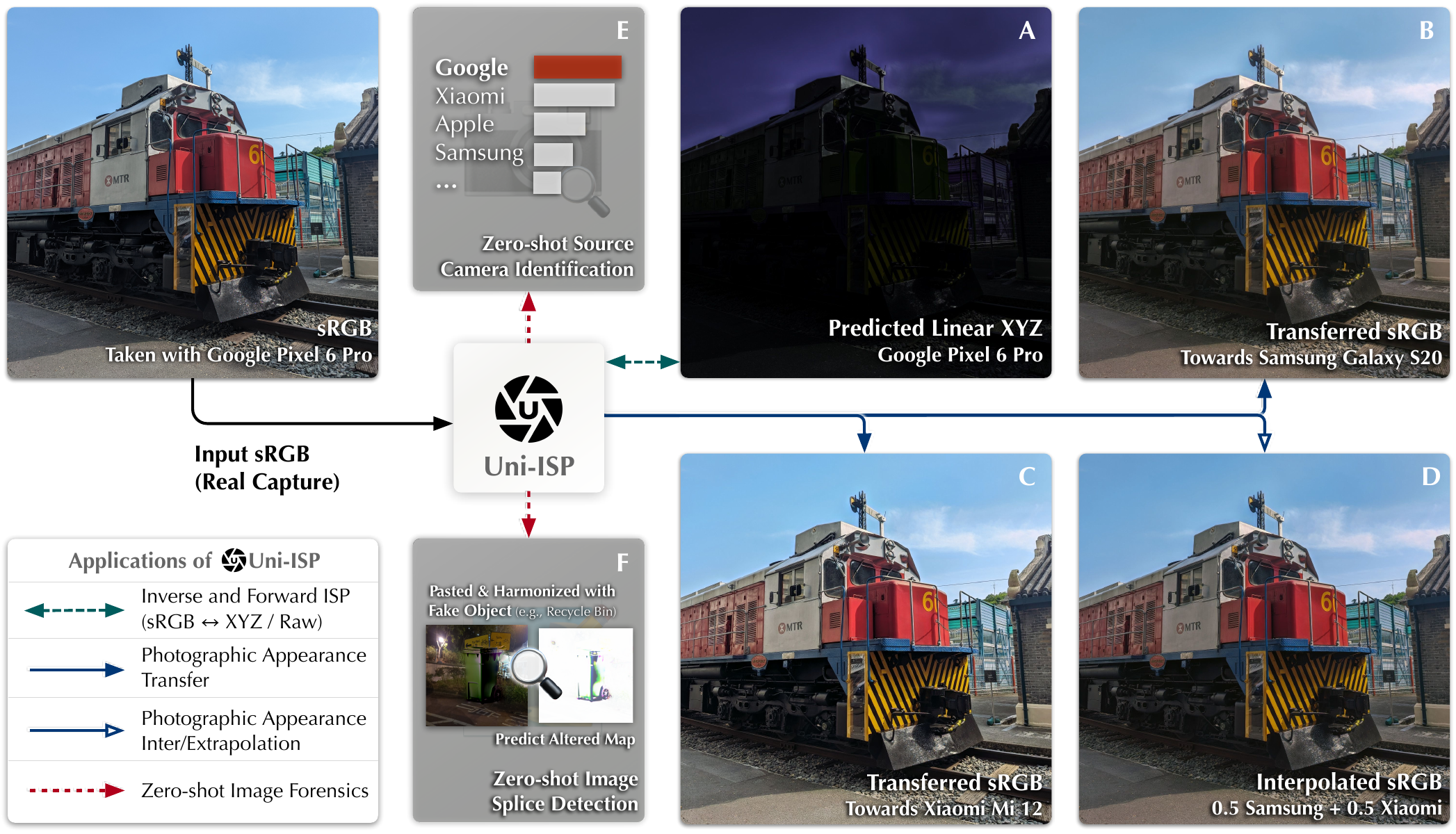}
   \caption{
    We propose \methodabbr, a model that learns the inverse and forward ISP behaviors of multiple mobile cameras simultaneously. By leveraging the shared characteristics across various camera ISPs, our method can achieve higher performance in inverse and forward ISP (A) compared to previously learned ISP methods tailored for only one camera separately. Meanwhile, the device-aware property of the \methodabbr enables new cross-camera ISP applications for a learned ISP model, including photographic appearance transfer (B and C), inter/extrapolation (D), and zero-shot image forensics (E and F). }
   \label{fig:teaser}
\end{figure*}

\begin{abstract}
Modern end-to-end image signal processors (ISPs) can learn complex mappings from RAW/XYZ data to sRGB (and vice versa), opening new possibilities in image processing. 
However, the growing diversity of camera models, particularly in mobile devices, renders the development of individual ISPs unsustainable due to their limited versatility and adaptability across varied camera systems.
In this paper, we introduce \textbf{Uni-ISP}, a novel pipeline that unifies ISP learning for diverse mobile cameras, delivering a highly accurate and adaptable processor. The core of Uni-ISP is leveraging device-aware embeddings through learning forward/inverse ISPs and its special training scheme. By doing so, Uni-ISP not only improves the performance of forward and inverse ISPs but also unlocks new applications previously inaccessible to conventional learned ISPs.
To support this work, we construct a real-world 4K dataset, \textbf{FiveCam}, comprising more than 2,400 pairs of sRGB-RAW images captured synchronously by five smartphone cameras.
Extensive experiments validate Uni-ISP's accuracy in learning forward and inverse ISPs (with improvements of +2.4dB/1.5dB PSNR), versatility in enabling new applications, and adaptability to new camera models.
\end{abstract}


\providecommand{\methodabbr}{Uni-ISP\xspace}
\providecommand{\methodfullname}{\textbf{D}evice-\textbf{a}ware \textbf{ISP}}
\providecommand{\methodfullnamebold}{\textbf{D}evice-\textbf{a}ware \textbf{ISP}}
\providecommand{\gfullname}{inverse ISP module\xspace}
\providecommand{\hfullname}{forward ISP module\xspace}

\section{Introduction}
\label{sec:intro}
Image Signal Processor (ISP) transforms raw image data captured by camera sensors into viewable formats such as sRGB, playing a pivotal role in determining the visual quality of photographs~\cite{karaimer2016software}. Through carefully designed ISPs, camera brands have developed distinctive photographic styles that appeal to diverse user preferences~\cite{deng2017image, souza2023metaisp}. For example, Apple's smartphone cameras are celebrated for their sharp and distinctive \textit{Apple feel}, while Leica cameras are esteemed for their glow and deep color tones, contributing to the iconic \textit{Leica look}.

Recently, neural networks have been used to approximate the entire ISP or a specific module, aka learned ISP, bringing two main benefits. 
1) \textit{Performance Enhancement}. The powerful representation capabilities of neural networks enable learned ISPs to perform challenging tasks, such as hallucinating detailed content in highlights and shadows~\cite{zou2023rawhdr}.
2) \textit{New Functionalities}. Learned ISPs introduce new functionalities, such as the inverse ISP~\cite{xing2021invertible, afifi2021cie}, which converts sRGB images back to RAW/XYZ space, offering greater flexibility and potential for raw-domain enhancement and further manipulation, such as deblurring~\cite{liang2020raw}, denoising~\cite{schwartz2018deepisp, chen2018learning, brooks2019unprocessing}, HDR photography~\cite{liu2020single, zou2023rawhdr}, etc. These innovations expand the applications and potential of learned ISPs. However, current methods design and train ISPs for individual camera models, which might limit the \textit{synergistic benefits} across different ISPs (see \Cref{sec:exp-inv-for-isp}). Moreover, versatile models have shown advantages in various fields in low-level vision~\cite{potlapalli2024promptir, li2022all}, high-level vision~\cite{kirillov2023segment}, and multi-modality~\cite{zeng2023x, xu2023versatile, wang2023all}.
As the number of camera models increases, individual learned ISPs may also lack the \textit{versatility} and \textit{adaptability} for widespread use, potentially making it unsustainable in the long term.

In this paper, our aim is to build a model toward unifying the learning of ISPs from different mobile cameras, which offers two direct advantages. First, it improves visual quality by leveraging the synergies among ISPs from multiple mobile cameras. 
Unified learning enables ISPs to understand the underlying commonalities and differences in the diverse data, leading to improved visual performance. 
Second, unified learning provides novel applications, surpassing the capabilities of existing learned ISPs limited to inverse and forward ISPs. Unified learning enables new uses such as photographic appearance transfer, interpolation, and extrapolation across diverse camera models. Additionally, as a bonus, a unified ISP model has the potential to support zero-shot image forensics based on the self-consistency of ISP behaviors, including image-level source camera identification and pixel-level image splice detection.

However, learning a device-aware ISP model for multiple cameras is far from trivial, presenting several challenges.
First, we observe that simply mixing training data from multiple cameras in current models~\cite{afifi2021cie, zamir2020cycleisp, xing2021invertible, kim2023paramisp} can not produce satisfactory performance.
Therefore, we propose \methodabbr, a novel unified ISP model for multiple cameras that contains several optimizable device-aware embeddings to learn the ISPs of different cameras. These device-aware embeddings enable the model to capture the specific characteristics tailored to individual devices, while the shared backbones capture underlying commonalities.

Second, current ISP datasets do not contain synchronized sRGB-Raw image pairs captured by more than three cameras.
Although these data are not necessary to learn the individual ISP, they are critical to learning a unified ISP that supports synergistic benefits across different camera models and enables the development of new applications.
To address this, we develop a synchronized camera array with five smartphones and construct a novel dataset, FiveCam, which consists of 2,464 synchronized high-quality sRGB-Raw-paired images with 4K spatial resolution. The captured dataset has a wide range of scenarios, from landscape to close-up, and contains different lighting conditions, including both indoor and outdoor settings during day and night.

Third, given the inevitable misalignment in synchronized photo pairs taken by different cameras, a robust alignment and training scheme is required.
To tackle this challenge, we first roughly align the images using optical-flow-based methods, which will introduce frequency bias in warped images. Then, we design a frequency bias correction (FBC) loss to mitigate texture blur. Additionally, we introduce the self-/cross-camera training schemes to facilitate applications across the same/different camera models.

With all of these three designs, our Uni-ISP can be applied to a wide range of image tasks (\Cref{fig:teaser}), such as photographic appearance transfer, interpolation, and extrapolation across diverse cameras. Users can apply the visual characteristics of one camera model to another, achieving unique aesthetic effects. 
It also facilitates zero-shot image forensic tasks by utilizing the self-consistency of these ISP behaviors, including source camera identification and image splice detection. Extensive experiments show that \methodabbr outperforms state-of-the-art methods with approximately 1.5dB PSNR in the inverse ISP and approximately 2.4dB PSNR in the forward ISP.

\providecommand{\methodabbr}{Uni-ISP\xspace}
\providecommand{\methodfullname}{\textbf{D}evice-\textbf{a}ware \textbf{ISP}}
\providecommand{\methodfullnamebold}{\textbf{D}evice-\textbf{a}ware \textbf{ISP}}
\providecommand{\gfullname}{inverse ISP module\xspace}
\providecommand{\hfullname}{forward ISP module\xspace}

\begin{figure*}[t]
  \centering
   \includegraphics[width=1\linewidth]{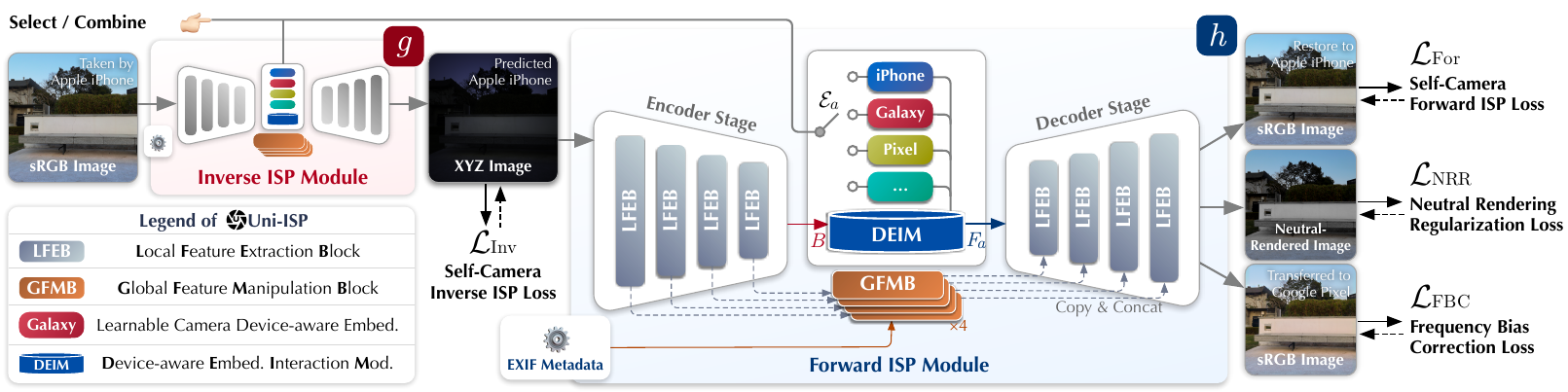}
   \caption{The model design of \methodabbr. \methodabbr contains two modules, the \gfullname $g$ and the \hfullname $h$. Both two modules share the same structure. For visual simplicity, we draw the \gfullname $g$ as a thumbnail, whose inner structure is the same as the \hfullname $h$. The device-aware embeddings are optimizable parameters and will be selected to interact with the bottleneck features via the DEIM during the training or inference.
   }
   \label{fig:model-design}
\end{figure*}

\section{Related Work}
\label{sec:related}
\paragraph{Learning Inverse and Forward ISP}
Training neural networks to learn the inverse and forward processes of certain ISPs has been explored in recent research works~\cite{afifi2021cie, zamir2020cycleisp, xing2021invertible}.
CycleISP~\cite{zamir2020cycleisp} maintains the self-contained property of inverse and forward ISP, where the cycle constraints are applied in addition to independent supervision from the RGB or the raw ground truth.
InvISP~\cite{xing2021invertible} achieves the inverse and forward ISP in one self-contained invertible network.
However, constraints that ensure the model's invertibility also limit the network's expression ability.  
ParamISP~\cite{kim2023paramisp} uses the EXIF data in the model design.
All existing methods ignore the variance of camera devices, overlooking the potential value of the commonalities across different camera devices in the learned ISP task. In contrast, our \methodabbr leverages these underlying commonalities by unifying the learning of multiple cameras.

\paragraph{Photorealistic Image Style Transfer}
Photorealistic image style transfer manipulates the aesthetics of an image, such as its color and tone, without distorting its original structure and content.
Non-learning-based photorealistic image style transfer is mainly handled through traditional image processing techniques~\cite{pitie2005n, pitie2007automated, nguyen2014illuminant}, including color grading~\cite{pitie2007automated},  gamut optimization~\cite{nguyen2014illuminant}, etc.
~\citet{gharbi2017deep} and ~\citet{xia2020joint} propose learning-based bilateral grids for photorealistic image style transfer.
Recent large generative models have also opened up new possibilities for photorealistic image style transfer~\cite{fu2022language, kwon2022clipstyler, liu2023name, brooks2023instructpix2pix}, but at the cost of huge computational resources and the risk of distorting the original content.
For efficient and high-fidelity photorealistic image style transfer, NeuralPreset~\cite{ke2023neural} is proposed for learning photorealistic image style transfer with a small memory footprint, which can be easily deployed on mobile devices. 
Unlike the aforementioned methods, our \methodabbr learns the transfer of style-like visual feeling produced by certain camera ISPs, the photographic appearance, which is device-dependent and physically faithful.

\paragraph{Image Forensics}
Image forensics focuses on validating and analyzing digital images to determine their origin by source camera identification or authenticity by splice detection. Source camera identification is highly related to the properties of the camera, which can be done by comparing the sensor pattern noise (SPN) and its main component photo response non-uniformity (PRNU) noise of a given image taken with an unknown camera and the reference image taken with known cameras.
~\citet{lukas2006digital} extract the PRNU using the discrete wavelet transform.
~\citet{chen2017camera} propose a residual network for source camera identification.
~\citet{hui2022source} propose a multi-scale feature fusion network and a corresponding two-stage training scheme for this task based on the guidance of PRNU.
Some methods are not based on PRNU but on the lens distortions and auto white balance (AWB) algorithms~\cite{deng2011source, bondi2016first}.
Unlike existing methods that require specific training, our method achieves zero-shot source camera identification by implicitly identifying the whole ISP behavior. Our model can also perform image splice detection by slightly modifying the inference process for source camera identification.

\providecommand{\methodabbr}{Uni-ISP\xspace}
\providecommand{\methodfullname}{\textbf{D}evice-\textbf{a}ware \textbf{ISP}}
\providecommand{\methodfullnamebold}{\textbf{D}evice-\textbf{a}ware \textbf{ISP}}
\providecommand{\gfullname}{inverse ISP module\xspace}
\providecommand{\hfullname}{forward ISP module\xspace}

\section{Method}
\label{sec:method}

\subsection{Overview}
First, we discuss the XYZ image format of the inverse and forward ISP tasks that our model will undertake. The XYZ images are device-independent measurements of radiances and learning XYZ images enjoys the same benefits as learning raw images. Therefore, in line with~\cite{afifi2021cie}, we opt for XYZ images processed from the raw images taken by the camera as the raw modality.
Specifically, the XYZ image is obtained from the real raw image by applying the early fixed stage of the ISP using the as-shot white balance, a fixed linear demosaicing algorithm, and the camera-to-XYZ matrix of the current device, without applying gamma tone mapping. In this setting, the XYZ images are linearly correlated to the raw images and they can be converted to each other without loss.

\Cref{fig:model-design} shows the overview of \methodabbr, which contains the \gfullname $g$ and \hfullname $h$.
Our model aims to be aware of various camera devices in the learned ISP tasks. 
Assume all images discussed here have dimensions $H \times W$ and $C$ channels. Given an sRGB image $I_a \in \mathbb{R}^{H\times W \times C}$ produced by the camera $a$, $\mathcal{E}_a \in \mathbb{R}^{D}$ represents the device-aware embedding for camera $a$ with size of $D$, the \gfullname $g$ learns to output the corresponding XYZ image $\hat{L}_a$ of camera $a$ given the input $I_a$:
\begin{equation}
    \hat{L}_a = g(I_a, \mathcal{E}_a),
    \label{eq:method-abstract-inv}
\end{equation}
and \hfullname $h$ learns to predict the $\hat{I}_a$ from the real XYZ image $L_a$:
\begin{equation}
    \hat{I}_a = h(L_a, \mathcal{E}_a).
    \label{eq:method-abstract-for}
\end{equation}
The $g(\cdot)$ and $h(\cdot)$  learn the inter-device general properties in ISP behaviors, while the device-aware embedding $\mathcal{E}_a$ focuses on device-specific intra-device properties of camera $a$. 
  
The above formulation allows us to concurrently learn ISP behaviors of multiple cameras $\{a, b, c, ..., z\}$ by training $g$ or $h$ alongside the device-aware embeddings $\{ \mathcal{E}_a, \mathcal{E}_b, \mathcal{E}_c, ..., \mathcal{E}_z \}$.

\subsection{Model Design}
\label{sec:method-detailed-design}
\begin{figure}[htbp]
    \centering
    \includegraphics[width=0.9\linewidth]{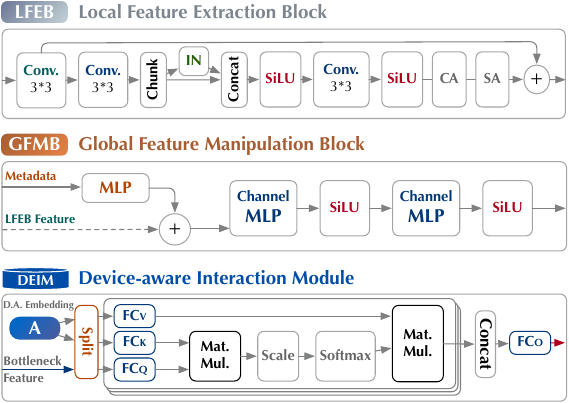}
    \caption{Illustration of the sub-components (LFEB, GFMB, and DEIM) inside Uni-ISP, where \textit{IN} refers to instance normalization, \textit{CA} stands for the channel-wise image attention layer, \textit{SA} stands for the spatial-wise image attention layer, and \textit{FC} represents the fully-connected linear layer.}
    \label{fig:subcomp}
\end{figure}

As depicted in \Cref{fig:model-design}, our \methodabbr utilizes the \gfullname $g$ and \hfullname $h$, each featuring an encoder-decoder architecture. Both modules incorporate Local Feature Extraction Blocks (LFEBs) for detailed local processing and Global Feature Manipulation Blocks (GFMBs) for broad image adjustments, mirroring the dual processing dynamics of real camera ISPs that manage both global operations like exposure compensation and color correction, and local tasks such as tone mapping and highlight recovery. The Device-aware Embedding Interaction Module (DEIM) adapts bottleneck features of the network into camera-specific features via an cross-attention mechanism with the learnable device-aware embeddings. 
\Cref{fig:subcomp} shows the detailed structure of the three components of our model design.

\paragraph{Local Feature Extraction Blocks} Each encoder and decoder stage in \methodabbr comprises four Local Feature Extraction Blocks (LFEBs). LFEBs in the encoder and decoder stages include max-pooling layers and upsampling layers respectively. Each LFEB contains multiple convolutional layers, activation layers, half instance normalization \cite{chen2021hinet}, and spatial/channel attention layers. Residual connections link LFEBs across the encoder and decoder stages.

\paragraph{Global Feature Manipulation Blocks} 
ISP behaviors can be affected by different parameters of exposure time and ISO, especially our dataset is collected with built-in auto-exposure strategies of each cameras. Different ISP parameters can lead to different global appearance of the photos in the same scene, and these parameters can be easily accessed form the JPEG or Raw files output by cameras.
Our GFMBs are designed to incorporate these parameters in the ISP learning process, which modifies the residual features from the encoder's LFEBs. These manipulated features are then relayed to corresponding LFEBs in the decoder stage. Used camera parameters (exposure, ISO, and f-number) are extracted from the EXIF metadata of each JPEG images produced by cameras.

\paragraph{Device-aware Embedding Interaction Module}
Positioned between the encoder and decoder stages, the Device-aware Embedding Interaction Modules (DEIM) enhance the model's ability to adapt to different camera devices by interacting with device-aware embeddings. Given a device-aware embedding $\mathcal{E}_a$, the DEIM applies an attention-based transform to the bottleneck features $B$ from the encoder stage and output $F_a$. This setup enables the model to adaptively learn the ISP behaviors of multiple cameras concurrently.

\begin{figure}[t]
    \centering
    \includegraphics[width=1\linewidth]{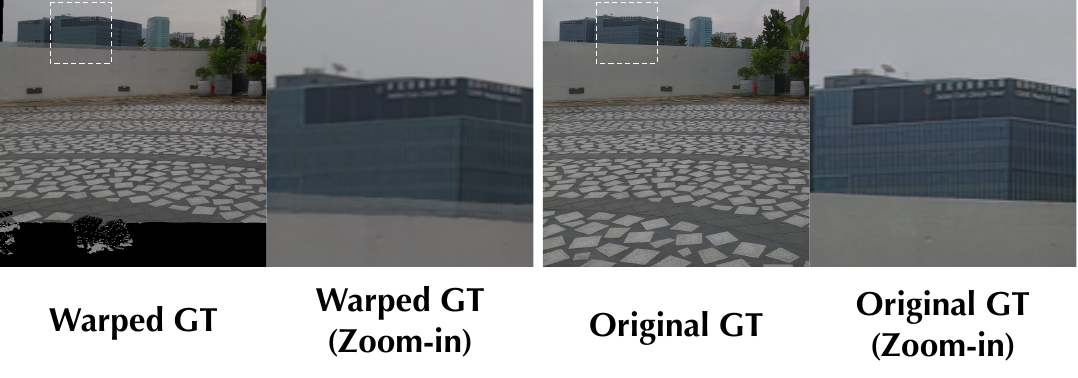}
    \caption{The illustration of frequency bias in dataset warped using optical flow method. The interpolation during the warping will make images look blurry compared to the original one, eliminating its high-frequency component. 
    }
    \label{fig:illustrateion-freq-bias}
\end{figure}

\subsection{Training Scheme}
\label{sec:method-training-scheme}
We craft a special training scheme for Uni-ISP, consisting of two types of training objectives: self-camera and cross-camera. Self-camera training ensures fair comparisons with prior single-camera ISP methods~\cite{kim2023paramisp,afifi2021cie,xing2021invertible}. Cross-camera training enables more applications like photographic appearance transfer across devices.

\subsubsection{Self-Camera Training Objective}
In the self-camera training objective, our \methodabbr learns the inverse and forward ISP behavior of each camera itself simultaneously. 
 
Given a sRGB image $I_a$ and corresponding XYZ image $L_a$ taken with camera $a$, the inverse ISP training objective is to minimize inverse ISP loss $\mathcal{L}_{\text{Inv}}$, aka the difference between $L_a$ and the predicted $\hat{L}_a$ given in~\Cref{eq:method-abstract-inv}:
\begin{equation}
\label{eq:obj-self-cam-inv}
     \mathcal{L}_{\text{Inv}} = \Vert L_a - \hat{L}_a \Vert_1.
\end{equation}
The forward ISP training objective requires the model to minimize the following forward ISP loss $\mathcal{L}_{\text{For}}$ that calculates differences between the ground truth $I_a$ and the $\hat{I}_a$ given in~\Cref{eq:method-abstract-for}:
\begin{equation}
     \mathcal{L}_{\text{For}} = \Vert I_a - \hat{I}_a \Vert_1.
    \label{eq:obj-self-cam-for}
\end{equation}

\subsubsection{Cross-Camera Training Objective}
The target photo in the cross-camera training objective is taken by a different camera than the one that captures the input photo.
Specifically, this training objective is applied on the \hfullname $h$ to handle the camera model transition. Assuming $I_a$ is taken by camera model $a$ and $I_b$ is taken by camera model $b$, the \gfullname $g$ of \methodabbr takes the $I_a$ as input and predicts the XYZ image $\hat{L}_a$, which is the same process as the one described in~\Cref{eq:method-abstract-inv}.
Then the \hfullname $h$ convert the predicted $\hat{L}_a$ into the sRGB image $\hat{I}_b$ of camera $b$:
\begin{equation}
    \hat{I}_b = h(\hat{L}_a, \mathcal{E}_b),
\end{equation}
where $\mathcal{E}_b$ indicate the device-aware embeddings of camera $b$. Here, $h$ uses $\hat{L}_a$ as input, which is different from the ground truth $L_a$ used in the self-camera training objective.
The cross-camera training objective is defined to minimize the distance between $\hat{I}_b$ and the ground truth $I_b$.

In the above definition, the input image $I_a$ and the output ground truth $I_b$ are not aligned since they are captured using different cameras. This makes pixel-level losses such as L1-loss fail to drive the cross-camera training objective that minimizes the distance between $\hat{I}_b$ and $I_b$.
Therefore, we first use the optical-flow-based method, RAFT~\cite{teed2020raft}, to warp our dataset for the cross-camera training objective. All sRGB images $I_b$ taken with camera $b$ will be warped to $I^w_b$, aligned with sRGB images $I_a$ taken with camera $a$. We apply this warping to every possible camera-to-camera sRGB pair in our dataset. Areas that cannot be aligned are marked as occlusions and masked during training.

However, although the optical-flow-based warping method effectively aligns these images, it also introduces a frequency bias in our dataset. As demonstrated in~\Cref{fig:illustrateion-freq-bias}, the warped image tends to have fewer high-frequency details compared to the image before warping. If we use the aligned images to train our model directly, it will unexpectedly learn to smooth the images in the cross-camera ISP tasks.

To address this problem, we propose the frequency bias correction (FBC) loss for the cross-camera training objective. 
The FBC loss mitigates the loss of high-frequency details due to optical-flow warping.
It combines a low-frequency L1 part to ensure learning the low-frequent part with the warped image $I_b^w$ (\textit{e.g.}, the color and tones), and a frequency-domain part $\mathcal{L}_{\text{Freq}}$ to preserve the original image’s sharpness from $I_b$.
The~\Cref{eq:app-cross-cam} shows the process of this task. The FBC loss can be written as:
\begin{equation}
    \mathcal{L}_{FBC} = \Vert f_{low}(\hat{I_b}) - f_{low}(I^w_b) \Vert_1 + \mathcal{L}_{Freq}(\hat{I_b}, I_b),
    \label{eq:fbcl}
\end{equation}
where $f_{low}$ is a low-pass filter and $\mathcal{L}_{Freq}$ is a loss in the frequency domain.
In our implementation, $f_{low}$ is a Gaussian filter with a kernel size of 5, and we adopt the focal frequency loss~\cite{jiang2021focal} as $\mathcal{L}_{Freq}$.

Finally, we use the FBC loss $\mathcal{L}_{FBC}$ to drive cross-camera training. Detailed ablation results on this loss function are presented in~\Cref{sec:exp-analysis}.

\begin{figure*}[!h]
    \centering
    \includegraphics[width=1\linewidth]{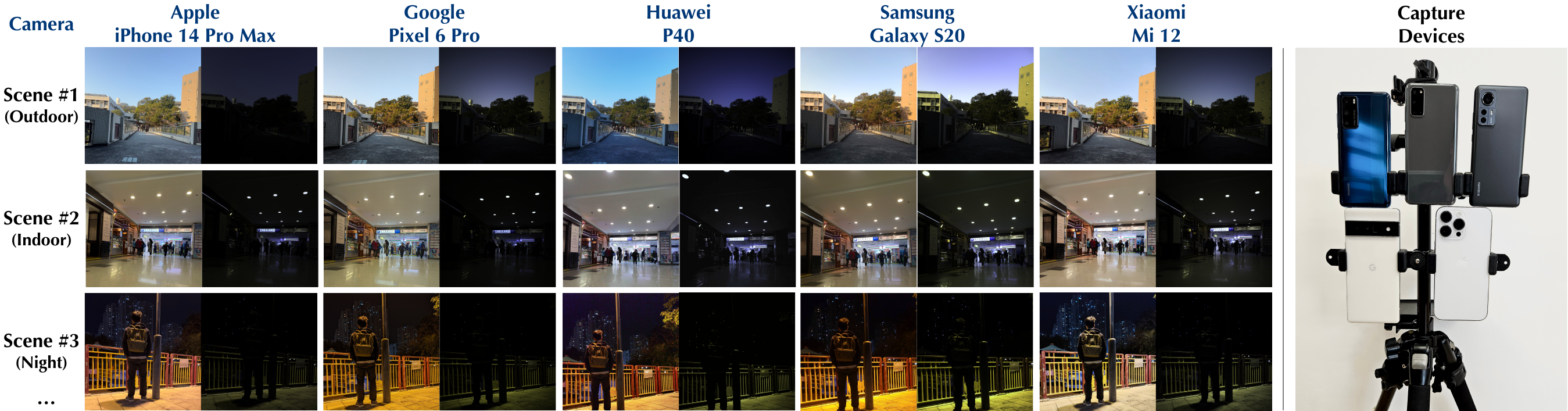}
    \caption{The preview of 3 scenes in our new dataset (left) and our capture devices (right). Each scene includes synchronized sRGB-Raw pairs of five smartphone cameras: Apple iPhone 14 Pro Max, Google Pixel 6 Pro, Huawei P40, Samsung Galaxy S20, and Xiaomi Mi 12. The raw images are visualized as XYZ images here, which can be converted back to raw without loss.
    }
    \label{fig:dataset-preview}
\end{figure*}

\subsubsection{Overall Loss}
\label{sec:method-overall-loss}

In summary, during the training of \methodabbr, we have three loss terms, the inverse ISP loss $\mathcal{L}_{\text{Inv}}$, the forward ISP loss $\mathcal{L}_{\text{For}}$, and the FBC loss $\mathcal{L}_{\text{FBC}}$ in total, and an additional regularization term $\mathcal{L}_{\text{NRR}}$:
\begin{equation}
    \mathcal{L} = \mathcal{L}_{\text{Inv}} + \mathcal{L}_{\text{For}} + \mathcal{L}_{\text{FBC}} + \lambda \mathcal{L}_{\text{NRR}},
    \label{eq:total-loss}
\end{equation}
where $\mathcal{L}_{\text{NRR}}$ is the additional neutral rendering regularization and $\lambda$ is the balance weight for it.
The neutral rendering regularization guides the model to learn a virtual camera that performs standard color conversion between the XYZ and sRGB color spaces when the device-aware embedding is given as a zero vector $\emptyset$:
\begin{equation}
    \mathcal{L}_{\text{NRR}} = \Vert s(I_a) - g(I_a, \emptyset) \Vert_1 + \Vert s^{-1}(L_a) - h(L_a, \emptyset) \Vert_1,
\end{equation}
where $s(\cdot)$ and $s^{-1}(\cdot)$ are the sRGB-XYZ and XYZ-sRGB color conversion, respectively. The neutral rendering regularization adds an anchor for users if they want to enhance or reduce the photographic appearance of a certain camera without inter/extrapolating with the device-ware embedding of another camera.

\subsection{Novel Dataset}
\label{sec:dataset}
Although existing sRGB-RAW datasets allow training models on both inverse and forward ISP tasks, there is still a need for datasets that contain synchronously captured sRGB-RAW pairs by multiple devices. Such datasets are essential for training models to handle cross-camera ISP tasks effectively, which are crucial for applications like photographic appearance transfer and inter/extrapolation.

To address this challenge, we collected a novel dataset named FiveCam, featuring synchronously captured sRGB-RAW pairs from five different camera models. This dataset encompasses 2,464 high-resolution (4K) raw and JPEG images representing approximately 500 diverse scenes. Cameras used in the FiveCam dataset include the Apple iPhone 14 Pro Max, Google Pixel 6 Pro, Huawei P40, Samsung Galaxy S20, and Xiaomi Mi 12. All cameras are synchronized using a programmed Bluetooth shutter to ensure consistent timing across all devices. 

A preview of the FiveCam dataset, illustrated in \Cref{fig:dataset-preview}, showcasing its three scenes along with the capture devices used. The scenes in our FiveCam dataset are richly varied, capturing both natural landscapes and urban environments under multiple lighting conditions, ranging from broad daylight to nighttime settings, and including both outdoor and indoor illumination.

Additionally, we have created an sRGB-XYZ version of this dataset, where raw images from all cameras are processed using as-shot white balance, a linear demosaicing algorithm, and converted to the standard camera-to-XYZ color space. XYZ images in this version retain their linearity, making them particularly beneficial for downstream tasks that require maintaining the linearity of raw images.

This dataset serves dual purposes: as a conventional raw image dataset for learning the inverse and forward ISP, and as a specialized resource for training models on cross-camera ISP tasks like photographic appearance transfer and inter/extrapolation. Please refer to the supplementary material provided for more detailed information about this dataset.

\providecommand{\methodabbr}{Uni-ISP\xspace}
\providecommand{\methodfullname}{\textbf{D}evice-\textbf{a}ware \textbf{ISP}}
\providecommand{\methodfullnamebold}{\textbf{D}evice-\textbf{a}ware \textbf{ISP}}
\providecommand{\gfullname}{inverse ISP module\xspace}
\providecommand{\hfullname}{forward ISP module\xspace}

\section{Applications}
\label{sec:exp}

\begin{figure}[]
    \centering
    \includegraphics[width=1\linewidth]{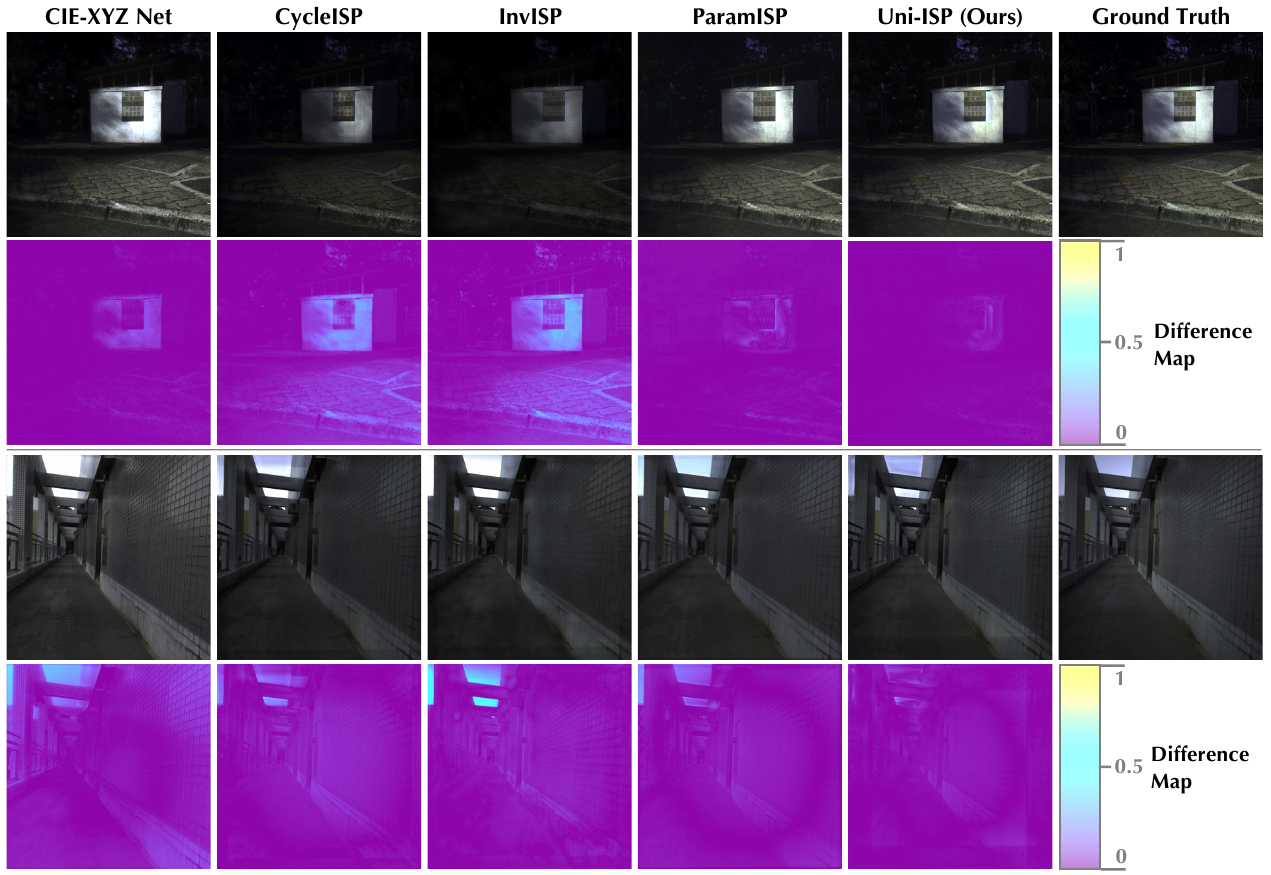}
    \caption{Two scenes that compare our \methodabbr with other methods in the task of inverse ISP. The difference maps between the prediction of each model and the ground truth are shown in the second and fourth rows. For better visualization, the XYZ images are adjusted with a 50\% increase in brightness to make the content easier to observe.}
    \label{fig:exp-self-isp-inverse}
\end{figure}

\begin{figure}[]
    \centering
    \includegraphics[width=1\linewidth]{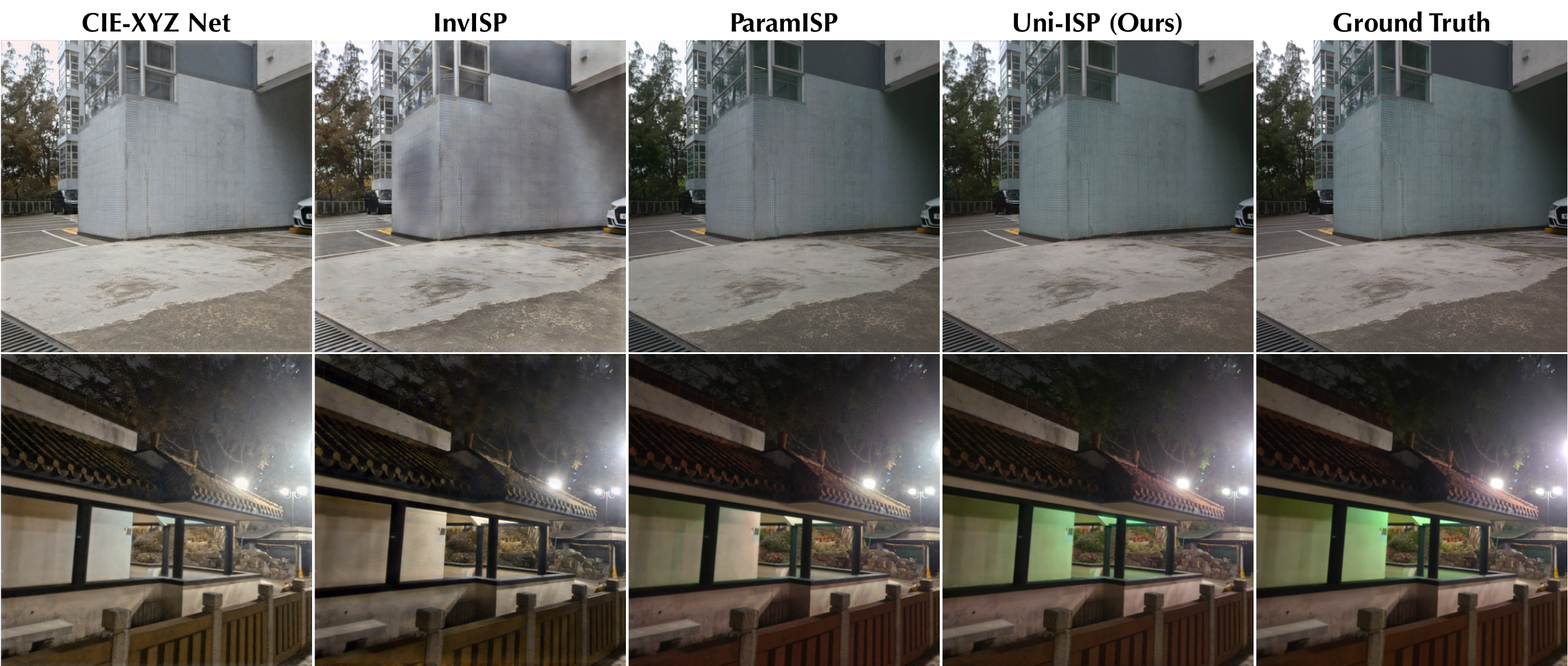}
    \caption{Two scenes that compare our \methodabbr with other methods in the task of forward ISP. The CycleISP~\cite{zamir2020cycleisp} is not included for a fair comparison since its forward module requires information from the ground truth RGB image.}
    \label{fig:exp-self-isp-forward}
\end{figure}

\begin{table}[]
\caption{Quantitative results of inverse and forward ISP tasks in the multi-camera mixed test set. All models are trained and tested with data from five different cameras. Please refer to~\Cref{sec:dataset} for detailed model names for each camera. To ensure fairness, the CycleISP~\cite{zamir2020cycleisp} is not included in the forward ISP test since it requires ground truth RGB input during the forward ISP process.}
\label{tab:self-isp-inv-for-isp}
\resizebox{\linewidth}{!}{
\begin{tabular}{@{}c|cccc@{}}
\toprule
\multirow{2}{*}{Method} & \multicolumn{2}{c}{Inverse ISP} & \multicolumn{2}{c}{Forward ISP} \\
                       & PSNR ($\uparrow$) & SSIM ($\uparrow$)  & PSNR  ($\uparrow$)  & SSIM ($\uparrow$)  \\
\midrule
Cycle ISP~\cite{zamir2020cycleisp} & 28.836         & 0.8632         & -              & -              \\
CIE-XYZ Net~\cite{afifi2021cie} & 24.990          & 0.7960          & 22.515         & 0.8750          \\
InvISP~\cite{xing2021invertible}   & 26.380          & 0.8042         & 21.644         & 0.8625         \\
ParamISP~\cite{kim2023paramisp} & 31.212         & 0.9180          & 26.739         & 0.9182         \\
\methodabbr (Ours)  & \textbf{32.699}         & \textbf{0.9396}         & \textbf{29.154}         & \textbf{0.9307}         \\ \bottomrule
\end{tabular}}
\end{table}

\begin{figure*}[htbp]
    \centering
    \includegraphics[width=0.9\linewidth]{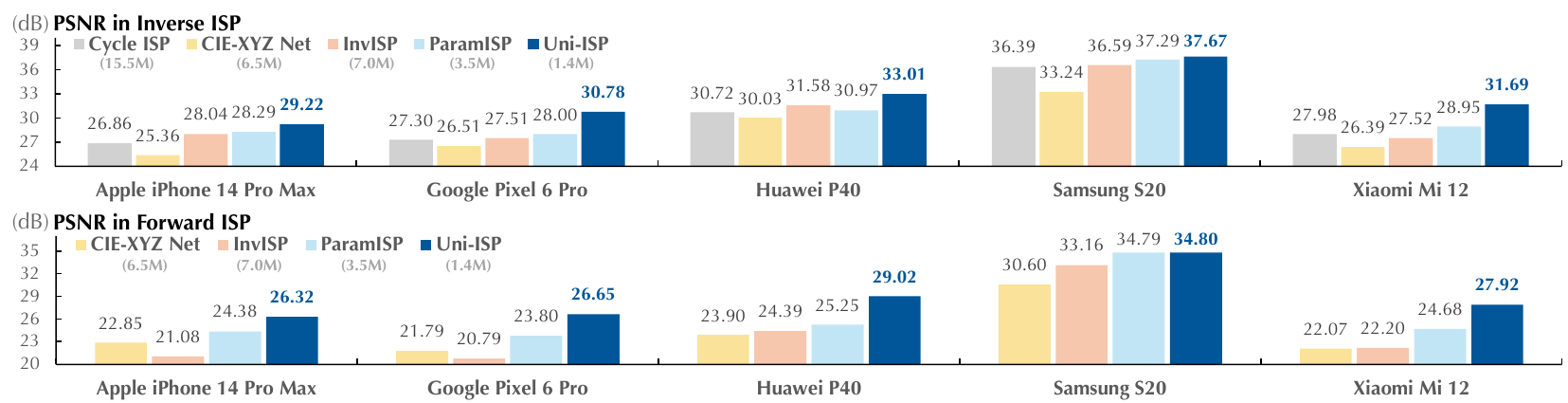}
    \caption{Quantitative results of inverse and forward ISP tasks in the single-camera test set. CycleISP~\cite{zamir2020cycleisp}, CIE-XYZ Net~\cite{afifi2021cie}, InvISP~\cite{xing2021invertible}, and ParamISP~\cite{kim2023paramisp} are trained for each camera model separately while our device-aware \methodabbr is trained on the mixed dataset. Numbers of total trained parameters for each method in this experiment are noted under the model name in gray. All models are tested with data from a single camera model (name mentioned in the horizontal axis), and the PSNR values are indicated in the vertical axis. For full numeric results in a table, please refer to the supplementary materials.}
    \label{fig:single-dev-comp-bar}
\end{figure*}

\begin{figure*}[htbp]
    \centering
    \includegraphics[width=0.9\linewidth]{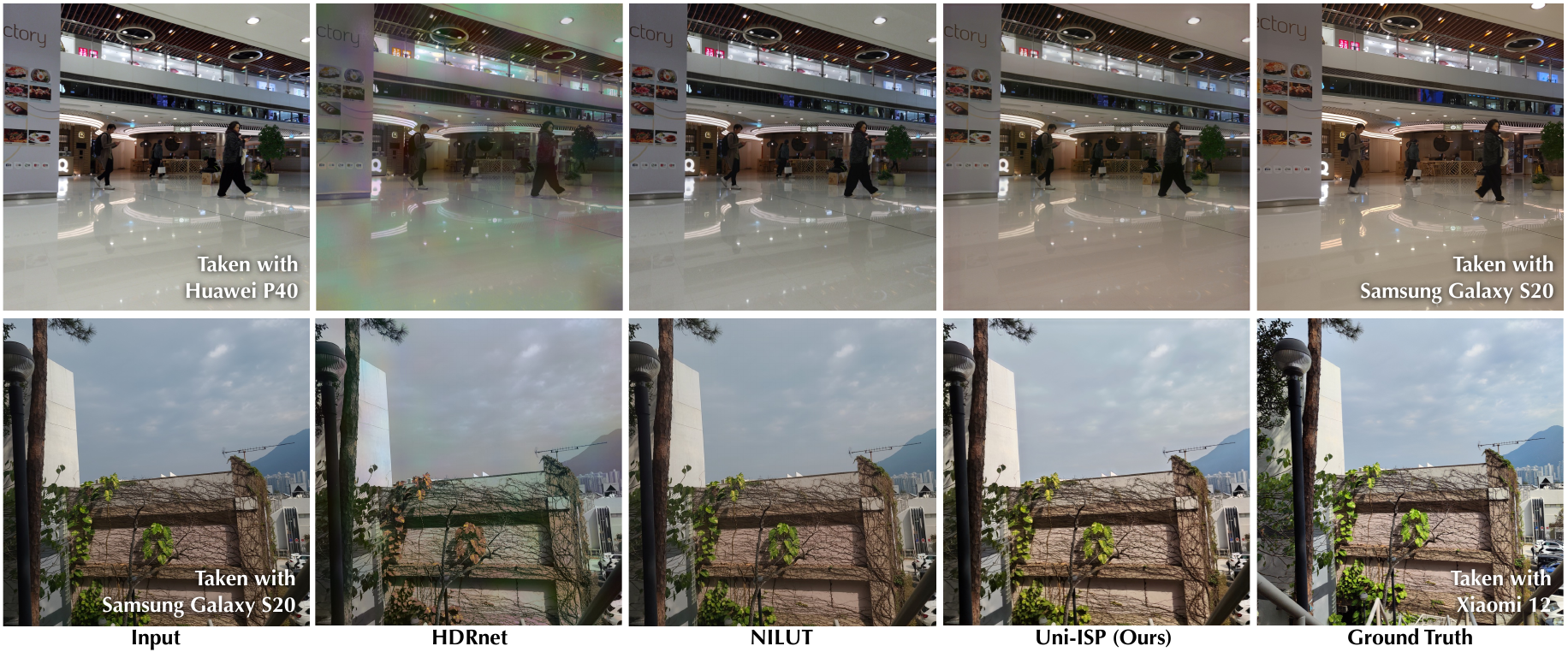}
    \caption{Comparison among \methodabbr, HDRnet~\cite{gharbi2017deep}, NILUT~\cite{conde2024nilut} in photographic appearance transfer. All three models are asked to transfer the photographic appearance to the camera that takes the ground truth photo. Our \methodabbr shows more accurate transfer results.}
    \label{fig:eval-transfer}
\end{figure*}

\begin{figure}[]
    \centering
    \includegraphics[width=1\linewidth]{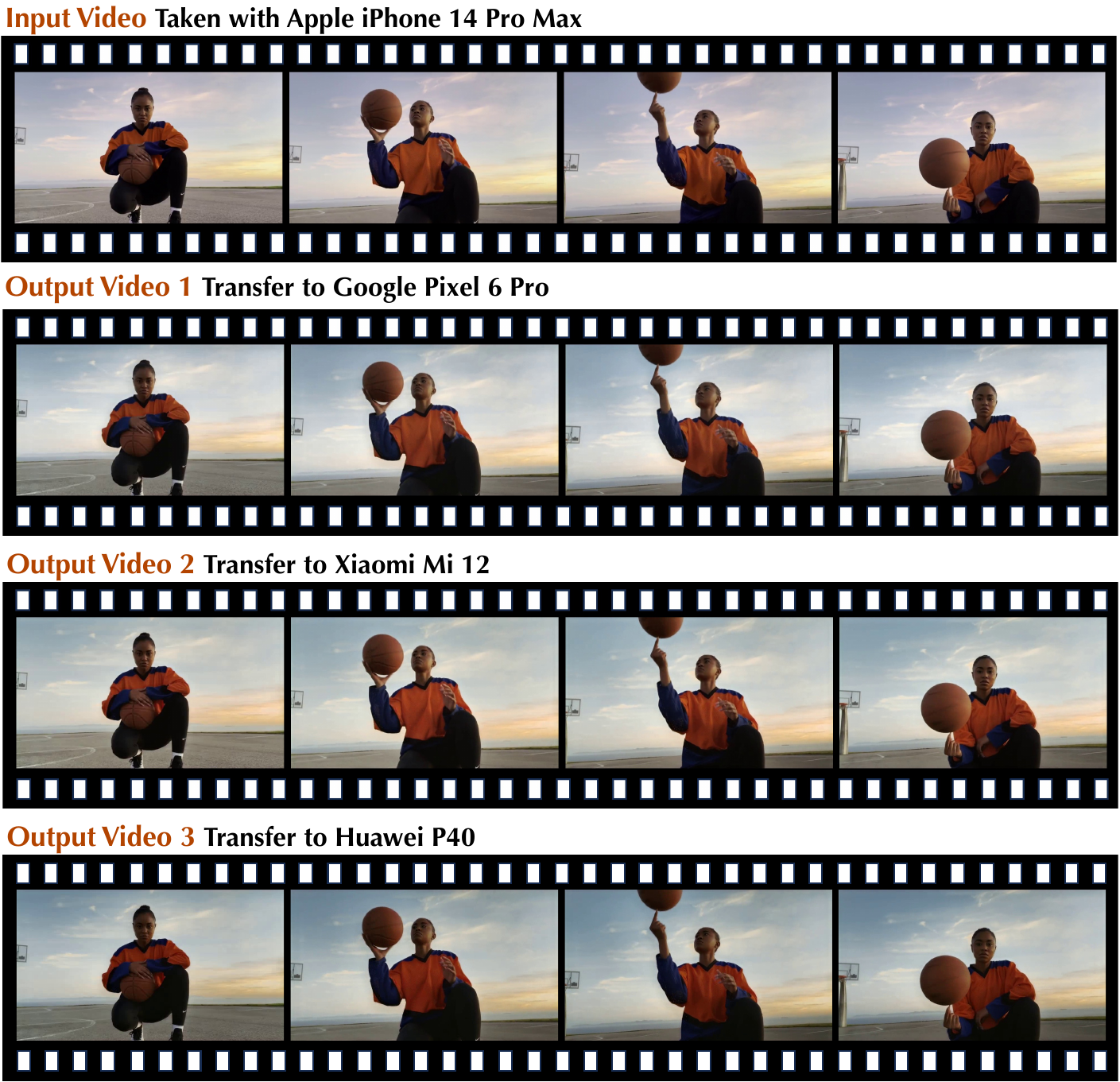}
    \caption{Results of photographic appearance transfer on videos with \methodabbr. We can see the consistent photographic appearance in the sky, even if our model does not have any temporal-specific design.}
    \label{fig:eval-video-trans}
\end{figure}
In this section, we conduct four experiments to validate \methodabbr on different applications, including 1) inverse and forward ISP, 2) photographic appearance transfer, 3) photographic appearance inter/extrapolation, and 4) zero-shot source camera forensics. We also conduct ablation studies and analytical experiments to verify the effectiveness of \methodabbr.

\subsection{Inverse and Forward ISP}
\label{sec:exp-inv-for-isp}
\paragraph{Settings} 
In the context of ISP learning, in addition to the forward ISP, inverse ISP enables raw-domain enhancements (e.g., HDR rendering, deblurring) from sRGB inputs when RAW data is unavailable, consistent with prior works \cite{kim2023paramisp,xing2021invertible,afifi2021cie}.
We compare \methodabbr with previous ISP learning methods in both inverse and forward ISP learning, including CycleISP~\cite{zamir2020cycleisp}, CIE-XYZNet~\cite{afifi2021cie}, InvISP~\cite{xing2021invertible}, and ParamISP~\cite{kim2023paramisp}.
All methods are retrained on the FiveCam dataset and we train two versions of baseline methods. The first version is trained on the mixed dataset, where sRGB-XYZ image pairs of all five cameras are used for training. The second version is trained on every single camera separately.
Our \methodabbr only trains a single version on the mixed dataset, with solely the self-camera training objective.

\paragraph{Results} For the first version, we show quantitative results in \Cref{tab:self-isp-inv-for-isp}, and qualitative results in~\Cref{fig:exp-self-isp-inverse} and~\Cref{fig:exp-self-isp-forward}. By observing the images and difference maps, our \methodabbr has less error compared to other methods in both two tasks, which is consistent with the numeric results shown in~\Cref{tab:self-isp-inv-for-isp}. 
For the second version, results in~\Cref{fig:single-dev-comp-bar} demonstrate our \methodabbr still outperforms all previous methods even when previous methods are trained on a single device separately, i.e., five models for each previous method.
In contrast, for our \methodabbr, we only need to train a single model.

\begin{table}[]
    \centering
    \caption{Quantitative results of photographic appearance transfer. The best value of each column is highlighted in bold.}
    \begin{tabular}{c|c|c}
        \toprule
        {Method} & {DISTS}($\downarrow$) & {PSNR}($\uparrow$) \\
        \midrule
        Learned Color Matrices & 0.1838 & 20.792 \\
        Learned 3D LUTs & 0.2027 & 20.853 \\
        Learned Bilateral Grids & 0.1971 & 20.625 \\
        HDRnet~\cite{gharbi2017deep} & 0.1722 & 20.863 \\
        NILUT~\cite{conde2024nilut} & 0.1434 & 21.390 \\
        \methodabbr (Ours) & \textbf{0.1392} & \textbf{24.237}  \\
        \bottomrule
    \end{tabular}
    \label{tab:exp-cam-ri-trans}
\end{table}

\subsection{Photographic Appearance Transfer}
\label{sec:exp-photo-appear-edit}

\paragraph{Settings} Photographic appearance transfer from the camera model $a$ to the camera model $b$ can be defined as:
\begin{equation}
\label{eq:app-cross-cam}
     I_b  = h(g(I_a, \mathcal{E}_a), \mathcal{E}_b),
\end{equation}
where $\mathcal{E}_a$ and $\mathcal{E}_b$ indicate two distinct device-aware embeddings of two camera models. This process uses the inverse module $g$ to inverse the image to the XYZ space and then uses forward module $h$ to generate the final image $I_b$.
Given the absence of previous work, we select a series of global and local color transform methods as baselines, including learnable global color transforms, learnable 3D LUTs (Look-Up Tables),  learnable bilateral grids, HDRnet~\cite{gharbi2017deep} and NILUT~\cite{conde2024nilut}. We input the camera label as the style code within HDRnet and NILUT for a fair comparison.
For the first three methods, we train 25 submodels to accommodate the various transfer mappings required by the 25 possible pairs of the five devices in FiveCam. Still, we only need to train one \methodabbr. We adopt PSNR and DISTS~\cite{ding2020image} for evaluation. 

\paragraph{Results} As shown in~\Cref{tab:exp-cam-ri-trans}, our method achieves the best results in terms of the two metrics, demonstrating the effectiveness of our method. 
We adopt DISTS to assess perceptual quality on unaligned original images, leveraging its robustness to structural differences.
PSNR is computed on warped ground truth images ($I_b^w$), excluding occlusion areas identified by the optical flow method, to measure pixel-level accuracy in aligned regions.
A visual comparison presented in \Cref{fig:eval-transfer} further illustrates that our \methodabbr achieves results much closer to the ground truth compared to previous methods such as HDRnet and NILUT, indicating the superior performance of our model. Moreover, our model has good temporal stability without a specific design. \Cref{fig:eval-video-trans} displays frames from the original iPhone video along with those transferred to emulate the photographic styles of Google and Xiaomi. Please refer to the supplementary materials for more details.

\subsection{Photographic Appearance Inter/Extrapolation}
\label{sec:eval-camri-interp}
\paragraph{Settings} In addition to transferring the photographic appearance, \methodabbr presents the ability to smoothly inter/extrapolate the photographic appearance between two cameras. Given the photo $I_a$ captured by the camera $a$, $I_a$ is firstly inverse to the XYZ image by $g$. Then, the XYZ image is encoded as the bottleneck feature and interacts separately with device-aware embeddings $\mathcal{E}_a$ and $\mathcal{E}_b$, producing intermediate features $F_a$ and $F_b$ right after DEIM. Then, the manipulated photo $I_{a+b}$ can be obtained by:
\begin{equation}
\label{eq:camrim-re-render}
     I_{a+b}= h_{decoder}((1-\alpha) {F}_a + \alpha {F}_b),
\end{equation}
where $h_{decoder}$ is the decoder part of $h$, $\alpha \in [0, 1]$ is the linear interpolation weight. For extrapolation, we relax the convex combination to the affine combination, where $\alpha$ is not limited in $[0, 1]$. 

\begin{figure*}[t]
    \centering
    \includegraphics[width=0.95\linewidth]{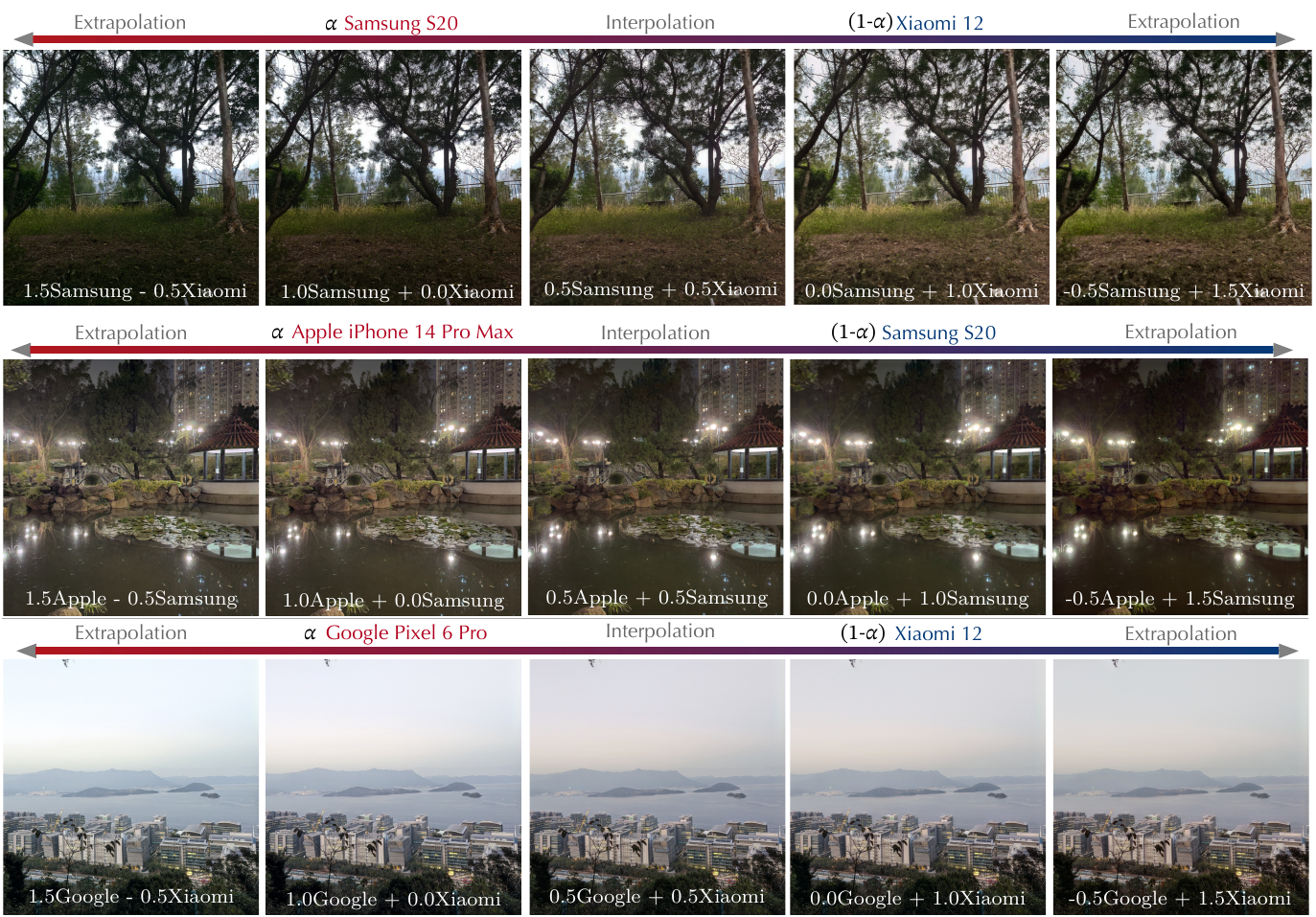}
    \caption{Photographic style interpolation and extrapolation between different camera models. We achieve this by adjusting the coefficient $\alpha$ in Eq.~\ref{eq:camrim-re-render}. Please refer to the electronic version on a bright display for better visualization.}
    \label{fig:eval-interp}
\end{figure*}

\paragraph{Results}
The results of interpolating and extrapolating photographic appearance are demonstrated in~\Cref{fig:eval-interp}. In the first scenario, the Samsung Galaxy S20 prefers a deeper shadow, and the extrapolation leads to a darker image. The second scenario is a night scene. Since Apple smartphones are famous for their night photography, our model also produces brighter results when we ask the model to be more ``Apple" where the lake and the trees become clearer.

\begin{figure}[]
    \centering
    \includegraphics[width=0.75\linewidth]{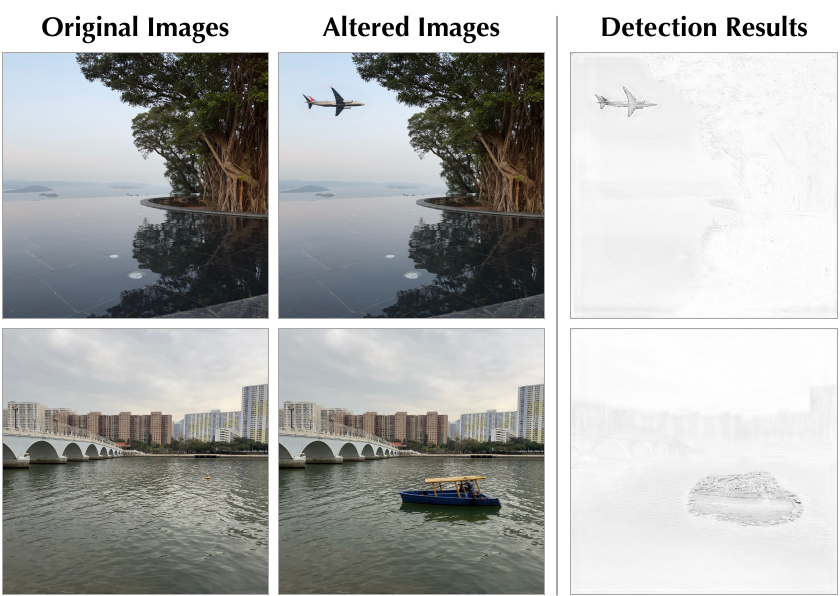}
    \caption{Image splice detection samples processed with generative region-wise editing for realistic tampering. Our method predicts a confidence map, with darker areas indicating likely alterations from the original capture.}
    \label{fig:exp-img-splice-detection}
\end{figure}

\begin{table}[htbp]
    \centering
    \caption{Source camera identification accuracy of CMIResNet~\cite{chen2017camera}, MSFFN~\cite{hui2022source}, and our \methodabbr.}
    \begin{tabular}{c|c}
        \toprule
        Method & Accuracy \\
        \midrule
        CMIResNet~\cite{chen2017camera} & 0.3833 \\
        MSFFN~\cite{hui2022source} &  0.4333 \\
        \methodabbr (Zero-shot) &  \textbf{0.8167} \\
        \bottomrule
    \end{tabular}
    \label{tab:exp-cam-src-id}
\end{table}

\subsection{Zero-shot Image Forensics} 
\label{sec:exp-img-forensics}
We perform two forensics tasks: source camera identification and image splice detection, \emph{without} training on these tasks. 

\subsubsection{Source Camera Identification}
\label{sec:app-cam-src-id}

\paragraph{Settings} 
During the inference of the source camera identification, Uni-ISP performs the inverse and forward ISP tasks multiple times with the device-aware embeddings of candidate camera models.
Then we measure the edit distance $D$ of each inference, and the predicted camera model is the one that leads to the minimal edit distance. 
We empirically choose structural similarity (SSIM)~\cite{wang2004image} as the metric to evaluate the editing distance $D$.
We compare our model with the camera model identification residual neural network (CMIResNet)~\cite{chen2017camera} and the multi-scale feature fusion network (MSFFN)~\cite{hui2022source}. 
These methods are trained on the same training set as the \methodabbr model, whose training objectives are classification tasks on the five cameras. For testing, we use 60 sRGB photos randomly chosen from a newly collected sRGB-only test set of unseen scenes captured by the same five cameras.

\paragraph{Results} 
\Cref{tab:exp-cam-src-id} shows the numeric results of the source camera identification experiments. Based on these results, our methods show superior performance over the conventional classification-based methods for source camera identification on our dataset, revealing a new possible perspective on this task.

\subsubsection{Image Splice Detection}
\paragraph{Settings}
Suppose that an image $I_a$ is taken with a camera $a$, but it is altered. We can predict a map to indicate the areas that are likely to be altered using the self-consistency of \methodabbr. This can be viewed as a pixel-level task in image forensics. We perform this task to show the versatility of our Uni-ISP.

\paragraph{Results}
\Cref{fig:exp-img-splice-detection} showcases \methodabbr's capability in image splice detection, where we introduce tampered objects into photos captured by an iPhone 14 Pro Max. These objects are seamlessly integrated using generative editing tools for realistic harmonization. The detection results are visualized as difference maps, with white regions indicating minimal differences (i.e., small editing distances) and darker regions highlighting areas likely altered from the original image. These results underscore \methodabbr's robust performance in unifying the learning of multiple camera ISPs for versatile forensic applications.

\subsection{Analysis}
\label{sec:exp-analysis}
\paragraph{Evaluation on HDR Rendering}
Learning the forward and inverse ISP benefits HDR rendering.
To validate the HDR rendering quality with currently learned inverse and forward ISP models, we synthesize multi-exposure raw image stacks with digital gains of $[1, 2.6, 4.2]$ on our test set, render them into sRGB images with a fixed ISP pipeline, and use the exposure fusion to produce the HDR tone-mapped images as ground truths.
Given an LDR sRGB image, we adopt models for inverse ISP to convert the LDR sRGB image into an XYZ image. Then, we synthesize a stack of frames with the same digital gains as the dataset and follow the same manner of the dataset to render them into an HDR tone-mapped image. 

\begin{figure}[t]
    \centering
    \includegraphics[width=0.96\linewidth]{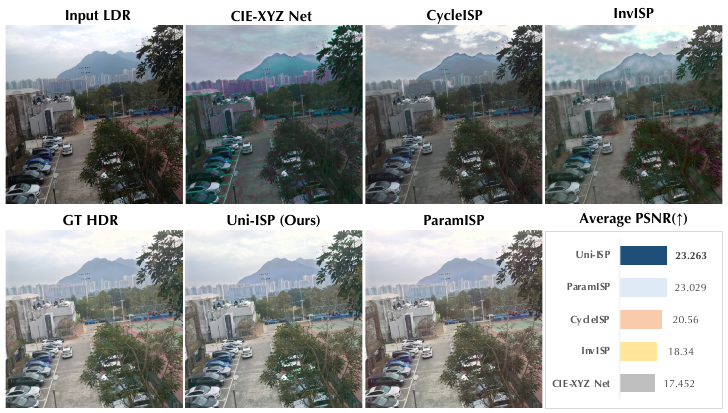}
    \caption{Visualization of the HDR rendering results and the average PSNR on the test set. Thanks to the accuracy in inverse ISP of \methodabbr, it produces the closest HDR rendering results to the ground truth.}
    \label{fig:hdr-rendering}
\end{figure}

\Cref{fig:hdr-rendering} shows the visualized results and the performance of each model in PSNR on the HDR rendering task. Consistent with the inverse ISP performance shown in~\Cref{tab:self-isp-inv-for-isp}, our method also achieves the best quality in the HDR rendering task.

\paragraph{Evaluation on Raw-Domain Deblurring}
The inverse ISP ability of~\methodabbr also makes it useful in raw-domain deblurring. We evaluate such ability of the~\methodabbr and other inverse ISP methods on the RealBlur dataset~\cite{rim2020real}. In this evaluation, we assume that we only have access to sRGB JPEG images, but we still want to conduct deblurring in the raw domain. We adopt the Stripformer~\cite{tsai2022stripformer} pretrained on raw images as the raw-domain deblurring network.

\begin{table}[h]
    \centering
    \caption{Numeric results of raw-domain deblurring using Stripformer~\cite{tsai2022stripformer} on the raw images produces by inverse ISP methods, with sRGB images given as input.}
    \begin{tabular}{c|c c}
        \toprule
        Inverse ISP Method & PSNR ($\uparrow$) & SSIM ($\uparrow$) \\
        \midrule
         CycleISP~\cite{zamir2020cycleisp} & 36.482 & 0.9382 \\
         CIE-XYZ Net~\cite{afifi2021cie} & 35.074 & 0.9619 \\
         InvISP~\cite{xing2021invertible} & 38.173 & 0.9767 \\
         ParamISP~\cite{kim2023paramisp} & 40.495 & 0.9784 \\
         Uni-ISP & 40.847 & 0.9799 \\
        \bottomrule
    \end{tabular}
    \label{tab:exp-deblurring}
\end{table}

\Cref{tab:exp-deblurring} shows the results of performing the raw-domain deblurring using inverse ISP methods when only sRGB images are available. Our Uni-ISP outperforms other inverse ISP methods in this raw-domain deblurring task, aligning with the inverse ISP accuracy we showed in~\Cref{sec:exp-inv-for-isp}.

\begin{figure}[t]
    \centering
    \includegraphics[width=0.8\linewidth]{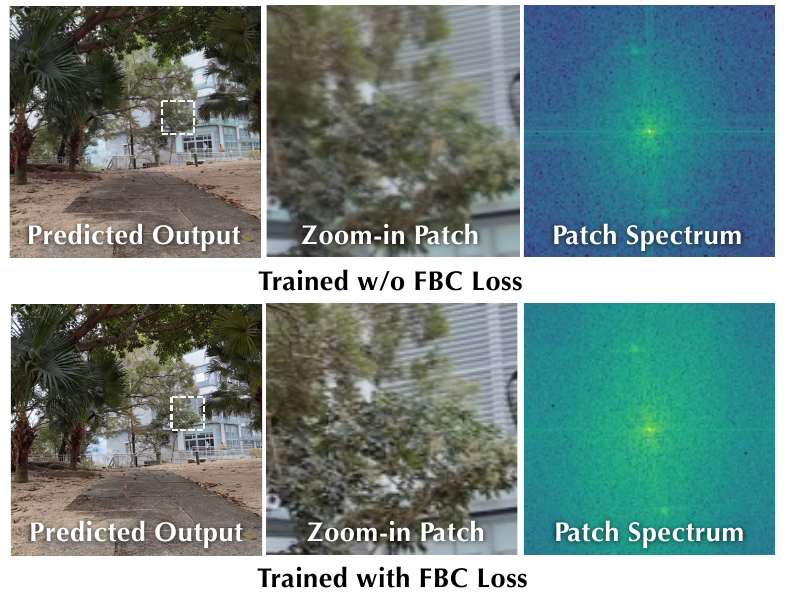}
    \caption{Ablation results on the FBC loss. With FBC loss, we effectively avoid our model to learn the frequency bias in our warped dataset. The spectrum visualization of the noted patch also demonstrates that the model trained with FBC loss produces a sharper image with stronger high-frequent components.}
    \label{fig:exp-ana-fbcl}
\end{figure}

\paragraph{Ablation Study on FBC Loss}
We perform ablation on FBC loss by replacing it with an L1 loss that minimizes differences between model predictions and warped images for training. As shown in the image and spectrum visualization of~\Cref{fig:exp-ana-fbcl}, the model trained with FBC loss produces clearer results than the one trained without FBC loss. Numeric results are presented in the supplementary material, which also demonstrates that the FBC loss effectively avoids frequency bias in the warped dataset.

\begin{table}[htbp]
\centering
    \caption{Ablation study on components in the model design of \methodabbr (GFMB, LFEB, and DEIM). The evaluation is conducted in self-camera ISP tasks. The best performance is noted in bold.}
    \label{tab:exp-ablation-components}
\begin{tabular}{@{}ccc|cc|cc@{}}
\toprule
\multicolumn{3}{c|}{Components} & \multicolumn{2}{c|}{Inverse ISP} & \multicolumn{2}{c}{Forward ISP} \\
GFMB     & LFEB     & DEIM     & PSNR           & SSIM           & PSNR           & SSIM           \\ \midrule
\xmark        & \xmark        & \xmark        & 26.181         & 0.8098         & 21.359         & 0.8658         \\ 
\xmark        & \cmark        & \xmark        & 28.782         & 0.8768         & 25.806         & 0.9090         \\ 
\cmark        & \cmark        & \xmark        & 29.010         &  0.8832        & 25.878         & 0.9094         \\ 
\xmark        & \xmark        & \cmark        & 30.940          & 0.9097         & 25.701         & 0.9102         \\
\xmark        & \cmark        & \cmark        & 32.320          & 0.9354         & 28.340          & 0.9245         \\
\cmark        & \cmark        & \cmark        & \textbf{32.699} & \textbf{0.9396} & \textbf{29.151} & \textbf{0.9306} \\
\bottomrule
\end{tabular}
\end{table}

\paragraph{Ablation Study on Components of \methodabbr}
We conduct ablation experiments on the LEFB, GFMB, and DEIM of \methodabbr. When the LFEB is disabled, we use an ordinary convolutional layer from vanilla U-Net as replacement. When the GFMB is disabled, it simply becomes a residual skip connection. When the DEIM is disabled, the bottleneck feature will be passed to the decoder stage directly from the encoder stage.
The results of the ablation experiments on two self-camera ISP tasks, which involve learning the inverse and forward ISP, are presented in~\Cref{tab:exp-ablation-components}. 

The LEFB extracts multi-scale features while the GFMB provides global manipulation based on the EXIF metadata. Therefore, both components contribute to improving the performance of \methodabbr. 
It is worth noting that the first row in~\Cref{tab:exp-ablation-components} shows a significant performance drop without the DEIM. This indicates that \methodabbr becomes a non-device-aware model as in all previous works, leading to a drastic performance drop in tasks involving multiple camera devices.

\begin{figure}[t]
    \centering
    \includegraphics[width=1\linewidth]{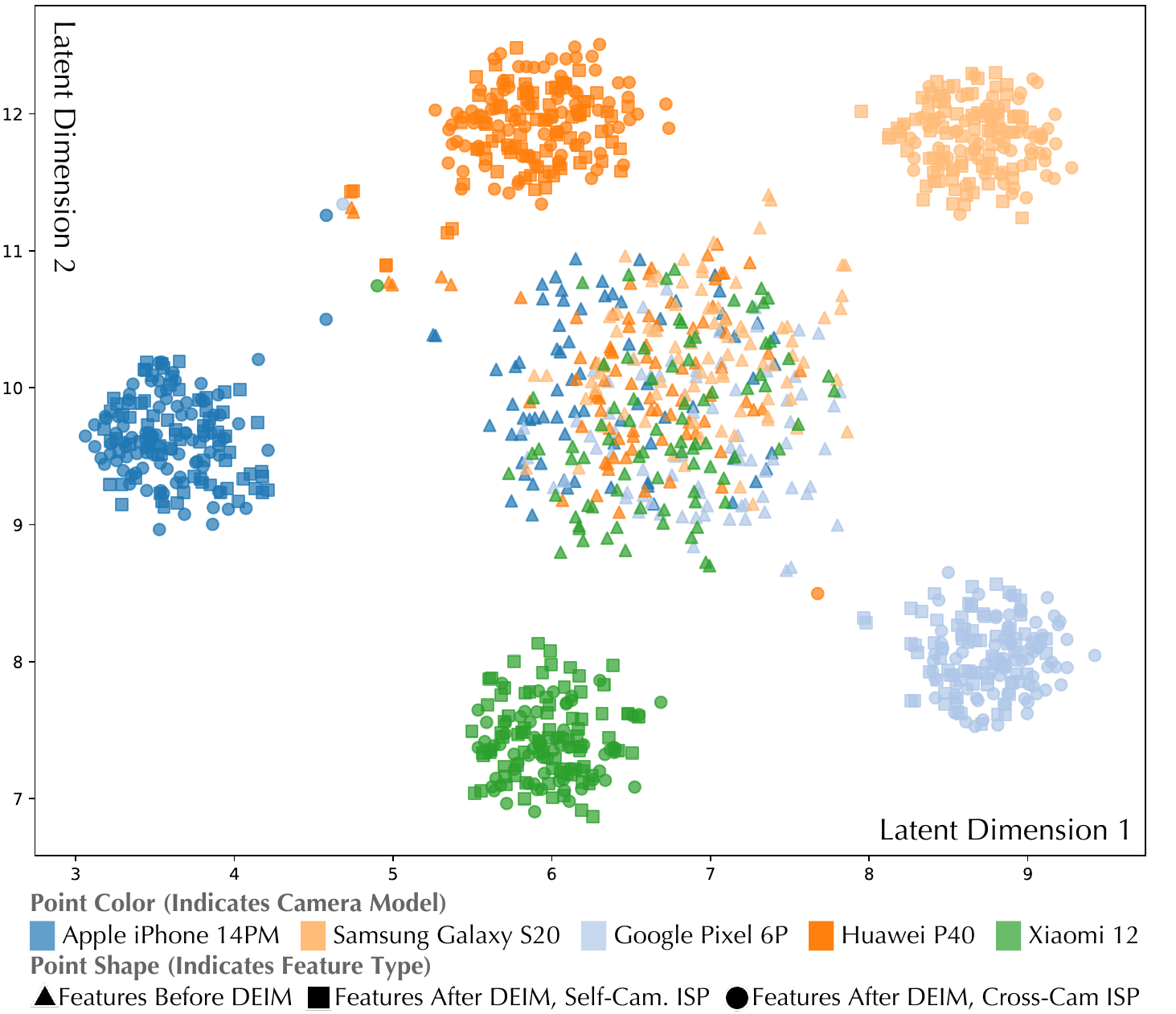}
    \caption{UMAP visualization of the internal features $B$ right before the DEIM and internal features $F_x$ ($x \in \{a, b, c, d, e\}$) right after the DEIM inside the \hfullname $h$ of \methodabbr. The DEIM lets corresponding device-aware embeddings $\mathcal{E}_x$ interact with the bottleneck features produced by the encoder stage of \hfullname ($h_{encoder}$), then produce the $F_x$. Features $F_x$ lie in a special manifold, where features in both self-camera and cross-camera tasks with the same target camera models are clustered together.}
    \label{fig:exp-ana-visual}
\end{figure}

\paragraph{Visualization of Internal Features}
To elucidate why \methodabbr effectively handles seamless inter/extrapolation of photographic appearances across multiple cameras, we visualize the internal feature distributions of our model. Specifically, we execute both self-camera tasks (inverse and forward ISP) and cross-camera tasks (photographic appearance transfer) using our test set. During these tests, we capture the \hfullname $h$'s intermediate bottleneck features $B$ before they are processed by the DEIM and $F_x$ for each camera $x \in \{a, b, c, d, e\}$ after they are processed by the DEIM. Here, $x \in \{a, b, c, d, e\}$ denotes the variable for the five cameras in our new dataset. 
The locations of $B$ and $F_x$ are noted in~\Cref{fig:model-design}.
We utilize UMAP~\cite{mcinnes2018umap} to project these high-dimensional features into a 2D space, as shown in \Cref{fig:exp-ana-visual}. This visualization clearly delineates the transformation dynamics of features, with distinct clusters emerging for each camera model. This clustering visually confirms that the process of interpolating or extrapolating photographic appearances mimics sampling from a well-defined manifold of camera ISP behaviors. Thanks to the DEIM, which transforms features from an indistinct subspace (points noted by triangles) into a clearly delineable manifold (points noted by circles and squares), our model facilitates smooth inter/extrapolation between the photographic appearances of different camera models, as depicted in \Cref{fig:eval-interp}.

\begin{table}[htbp]
    \centering
    \caption{Few-shot extension experiments for learning unseen cameras'  inverse and forward ISP. Images are taken from the S7 ISP dataset~\cite{schwartz2018deepisp} and our additional extension sub-set, covering mobile cameras including Samsung Galaxy S7, Apple iPhone 15 Pro, Apple iPhone 8 Plus, and SONY Xperia 5 II. The best performance for each camera is noted in \textbf{bold}.}
    \resizebox{\linewidth}{!}{
    \begin{tabular}{c|c|c c|c c}
        \toprule
        \multirow{2}{*}{Camera} & \multirow{2}{*}{Method} & \multicolumn{2}{c|}{Inverse ISP} & \multicolumn{2}{c}{Forward ISP} \\
        &  & PSNR ($\uparrow$) & SSIM ($\uparrow$)  & PSNR  ($\uparrow$)  & SSIM ($\uparrow$)  \\
        \midrule
\multirow{2}{*}{S7} & ParamISP & 23.907 & 0.8305 & 21.439           & 0.8489          \\
& \methodabbr & \textbf{25.307} & \textbf{0.8597} & \textbf{22.798}  & \textbf{0.8590} \\
\midrule
\multirow{2}{*}{15Pro} & ParamISP & 24.281 & 0.5122 & 16.377 & 0.6266          \\
& \methodabbr & \textbf{25.301} & \textbf{0.7380} & \textbf{23.086}  & \textbf{0.8429} \\
\midrule
\multirow{2}{*}{8Plus} & ParamISP & 14.952 & 0.3494 & 17.238 & 0.7046          \\
& \methodabbr & \textbf{24.922} & \textbf{0.8396} & \textbf{23.354}  & \textbf{0.9251} \\
\midrule
\multirow{2}{*}{Xperia 5 II} & ParamISP & 13.802 & 0.1360 & 16.619 & 0.3985          \\
& \methodabbr & \textbf{19.848} & \textbf{0.6428} & \textbf{20.480} & \textbf{0.518} \\

\bottomrule
    \end{tabular}}
    \label{tab:exp-extension}
\end{table}

\paragraph{Few-shot Extension Ability}
The proposed \methodabbr has the extension ability to perform well in the inverse and forward ISP learning of that new camera by only training a new device-aware embedding while keeping the rest of the model frozen.
This requires a few sRGB-Raw pair samples captured by the target unseen camera. Adding a new camera requires collecting paired data for retraining, as modifying EXIF metadata alone is not sufficient to mimic the behavior of a new ISP.

We conduct the few-shot extension experiments on the S7 ISP dataset~\cite{schwartz2018deepisp} and an additional small-scale dataset collected by ourselves using three additional cameras (iPhone 15 Pro, iPhone 8 Plus, and SONY Xperia 5 II) to validate the few-shot extension ability of \methodabbr. 
During training, we load the weights of the previously trained model in \Cref{sec:exp-inv-for-isp} and only optimize the newly created device-aware embedding for this unseen camera.
For comparison, we select ParamISP~\cite{kim2023paramisp} since it outperforms previous methods in most cases.
Unlike our method, which only requires training new device-aware embeddings for unseen cameras, ParamISP requires full retraining to adapt to a new camera. 
Simply modifying the EXIF input of ParamISP is not enough to mimic an unseen camera's ISP behavior.
We randomly select 10 scenes taken by each camera for training and use the other 100 scenes for testing. All models are trained for 50 epochs for convergence. Tests are conducted at the checkpoint of the 50th epoch. \Cref{tab:exp-extension} shows the results of few-shot extension experiments to learn the inverse and forward ISP of unseen camera models. As we can see, our \methodabbr outperforms ParamISP~\cite{kim2023paramisp} when only few training samples are provided, which demonstrates the value of this extension ability in extreme scenarios with inadequate training datasets.

\providecommand{\methodabbr}{Uni-ISP\xspace}
\providecommand{\methodfullname}{\textbf{D}evice-\textbf{a}ware \textbf{ISP}}
\providecommand{\methodfullnamebold}{\textbf{D}evice-\textbf{a}ware \textbf{ISP}}
\providecommand{\gfullname}{inverse ISP module\xspace}
\providecommand{\hfullname}{forward ISP module\xspace}

\section{Conclusion}
In this study, we presented the \methodabbr, which can concurrently learn inverse and forward ISP behaviors of multiple cameras. The \methodabbr harnesses the commonalities and distinctions across different cameras, enhancing the performance of ISP tasks beyond that of models trained on single-camera or mixed datasets. Furthermore, \methodabbr introduces novel applications for learned ISPs, including photographic appearance editing and zero-shot camera source forensics.

In future work, we plan to explore the interaction between device-specific components and the backbone network by introducing more domain knowledge from ISP designs. In addition, considering the complexity of inverse ISP tasks, which can often be highly ill-posed, investigating the use of generative priors to maintain image fidelity presents an intriguing avenue for future research.

\bibliographystyle{IEEEtranN}
\bibliography{main}

\end{document}